\documentclass[sigconf]{acmart}
\usepackage{algorithm}
\usepackage{algpseudocode}
\usepackage{caption}
\usepackage{subcaption}

\usepackage{placeins}

\usepackage{tcolorbox}
\usepackage{xcolor}

\definecolor{identitycolor}{RGB}{0,102,204}   
\definecolor{rulescolor}{RGB}{204,0,0}        
\definecolor{candidatecolor}{RGB}{0,153,0}    
\definecolor{utterancecolor}{RGB}{153,51,153} 

\usepackage{colortbl}

\definecolor{lightred}{rgb}{1.0,0.8,0.8}
\definecolor{mediumred}{rgb}{1.0,0.6,0.6}
\definecolor{darkred}{rgb}{1.0,0.4,0.4}
\definecolor{lightgreen}{rgb}{0.8,1.0,0.8}
\definecolor{mediumgreen}{rgb}{0.6,1.0,0.6}
\definecolor{darkgreen}{rgb}{0.4,1.0,0.4}

\AtBeginDocument{%
  }

\setcopyright{acmlicensed}
\copyrightyear{2025}
\acmYear{2025}
\acmDOI{XXXXXXX.XXXXXXX}
\acmConference[Prompt Optimization KDD '25]{First International KDD Workshop on Prompt Optimization, 2025}{Aug 04, 2025}{Toronto, ON, Canada}

\begin{document}

\title{How and Where to Translate? The Impact of Translation Strategies in Cross-lingual LLM Prompting}


\author{Aman Gupta}
\affiliation{%
  \institution{Amazon}
  \city{San Francisco}
  \country{USA}
}
\email{amangta@amazon.com}

\author{Yingying Zhuang}
\affiliation{%
  \institution{Amazon}
  \city{San Francisco}
  \country{USA}
}
\email{yyzhuang@amazon.com}

\author{Zhou Yu}
\affiliation{%
  \institution{Amazon}
  \city{Seattle}
  \country{USA}
}
\email{amznzya@amazon.com}

\author{Ziji Zhang}
\affiliation{%
  \institution{Amazon}
  \city{Seattle}
  \country{USA}
}
\email{czhangzi@amazon.com}

\author{Anurag Beniwal}
\affiliation{%
 \institution{Amazon}
  \city{San Francisco}
  \country{USA}
 }
 \email{beanurag@amazon.com}

\renewcommand{\shortauthors}{Gupta et al.}

\begin{abstract}
Despite advances in the multilingual capabilities of Large Language Models (LLMs), their performance varies substantially across different languages and tasks. In multilingual retrieval-augmented generation (RAG)-based systems, knowledge bases (KB) are often shared from high-resource languages (such as English) to low-resource ones, resulting in retrieved information from the KB being in a different language than the rest of the context. In such scenarios, two common practices are pre-translation to create a mono-lingual prompt and cross-lingual prompting for direct inference. However, the impact of these choices remains unclear. In this paper, we systematically evaluate the impact of different prompt translation strategies for classification tasks with RAG-enhanced LLMs in multilingual systems. Experimental results show that an optimized prompting strategy can significantly improve knowledge sharing across languages, therefore improve the performance on the downstream classification task. The findings advocate for a broader utilization of multilingual resource sharing and cross-lingual prompt optimization for non-English languages, especially the low-resource ones. 
\end{abstract}




\maketitle

\section{Introduction}

Text classification has always been an active Natural Language Processing (NLP) task with applicability to a wide range of real-world problems, including sentiment analysis, topic labeling, conversation dialogue state tracking, etc. It aims to choose one or multiple labels from a list of pre-defined categories given some informative texts. As Large Language Models (LLMs) have emerged as a powerful tool for language understanding and generation\cite{DialogSumm, zhuang-etal-2021-weakly}, LLM-based classification has become an active area of research and has achieved remarkable advancements \cite{zhang2024pushing, yin2024crisissense, bucher2024fine}. In dialogue systems and conversation AIs, a Retrieval-Augmented Generation (RAG) component is often deployed to generate a list of probable candidates for the task and a LLM is then used as a re-ranker to choose the final label from the list \cite{ampazis2024improving}. A multilingual retriever has become a common practice where the retrieved information and candidate lists are in a different language from the rest contexts and the label. Figure \ref{fig:example-prompt} demonstrates such an example for user intent classification where given a user utterance in French, a multilingual retriever will identify texts in English in the Knowledge Base (KB) with similar semantics and use their corresponding user intents as the candidate list. Therefore the candidate list is in English while the user utterance is in French, resulting in a cross-lingual prompt. 

One common strategy is to pass this prompt directly to a LLM with multilingual capability to predict the final user intent. Many of today's frontier LLMs, such as Anthropic’s Claude 3 \cite{claude3} and Cohere’s Command A \cite{cohere2025commandaenterprisereadylarge} are focused to excel in the multilingual settings to support such business needs. However, the majority of the state-of-the-art LLMs cover only a small percentage of the world’s spoken languages and heavily favor those with abundant resources such as English \cite{nicholas2023lost, zhao2024llama, naous2024havingbeerprayermeasuring, li2024quantifying}.  This creates a gap in the availability of models that can be applied for direct inference with cross-lingual prompts, specially when the prompt contains minority languages, as well as a bias in performance towards English and other high-resource languages compared to the lower resourced ones. 

The second option is pre-translation, where parts of the prompt are translated to ensure monolingualism (often English) before applying a LLM as the re-ranker \cite{ahuja2023mega, shi2022language}. This removes the multilingual dependency of the LLM therefore ensures the optimal English performance is utilized. However, as translation may contain errors or terms native speakers don't actually use, or fail to account for the contexts of the local language, this strategy introduces complexities and risks of information loss \cite{nicholas2023lost}. 

In this study, we answer the specific question of \textit{which components of the input prompt should be pre-translated and into which language to achieve optimal performance}. We systematically evaluate six translation strategies across prompt components—from keeping all components in English to selectively translating candidates, identity statements, rules, or all components to the source language. We perform experiments on French and Hindi as the source languages as they respectively represent the high-resource and low-resource scenarios, with five diverse language models on the task of dialogue intent detection. Our performance results suggest that while an optimal pre-translated prompt can lead to improved performance, the impact of pre-translation varies significantly based on the model choice and the nature of the source language.

\section{Related Work}
Various studies have proposed to use selective pre-translation in multi-lingual settings, where only specific parts of the prompt is pre-translated to optimize performance. For example, \cite{liu2025translationneedstudysolving} demonstrated that translating only the context to English outperforms direct-inference in summarization and NLI tasks, while \cite{ahuja2023mega} showed that translating only the few-shot examples while keeping the rest of the prompt in the source language yielded optimal results for low-resource languages while does not make significant impact for high-resource languages compared to monolingual prompting. Furthermore, \cite{kim2024crosslingualqakeyunlocking} proposed a cross-lingual prompting method that translates only the question and answer parts for the QA task to optimize in-context learning. One common theme from all these studies is \textbf{how and where to translate} in cross-lingual prompting is highly dependent on 1) the task, 2) the language setting (high vs low resource), 3) translation quality, 4) the tokenizer, 5) and finally the LLM deployed for inference. This calls for the need to systematically study the effectiveness of various prompting strategies across different tasks and LLMs. 

\cite{mondshine-etal-2025-beyond} is aimed to address this need where they evaluate the impact of different pre-translation strategies across a range of tasks, including NLI, QA, NER, and Summarization, in order to provide general guidelines for choosing optimal strategies in various multilingual settings. However, they did not study LLM-based classification tasks where the prompt contains a list of probable candidates and the LLM is prompted to choose the final answer from the list based on the contexts provided in the prompt. In our work we specifically fill this gap for multilingual classification tasks, motivated by real-world applications where information is often retrieved from a multilingual KB. The ability
to share knowledge in multiple languages is a fundamental component to ensure the development of Conversation AIs systems to reach a broader range of communities, and we hope our work spurs research to foster inclusivity, accessibility and collaboration for both businesses and individuals.

\section{Methodology}

In this section, we outline the experimental framework we used to evaluate the performance impact of translating prompts into different languages. We begin by presenting the overall task description in Section \ref{task}, followed by a brief overview of the dataset in Section \ref{sec:dataset}. Section \ref{prompts} describes the different prompts we used and, Section \ref{models} details our choice of models for evaluation.

\subsection{Task Description}\label{task}
We assess the results on the task of intent classification in dialogue systems, where the goal is to detect the user's intent based on their utterances. We follow the two-step RAG setup for intent detection\cite{arora-etal-2024-intent, liu-etal-2024-lara,  zhang2025reicragenhancedintentclassification}, where a retrieval model first retrieves the list of candidate intents and then a LLM picks the final intent from the candidates. For the purpose of this study we focus on prompts for the LLM to pick the final intent from the candidates.  We enforce full recall from the retrieval model that is, the correct intent is always present in the candidates. Overall, the prompt contains i) an \textit{identity statement (\textbf{I})} which describes the high level idea of the task, ii) \textit{task rules (\textbf{R})} which present instructions such as how to work on the task and what output format to follow, iii) a \textit{candidate list (\textbf{C})} of the candidate intents and iv) the \textit{speaker utterance (\textbf{U})} for which we need to detect their intent. Figure \ref{fig:example-prompt} shows the prompt that we use for the task. We use accuracy as the primary performance evaluation metric. For each instance, we check if only the correct intent (verbatim) is present and no other labels are present in the model response. If these conditions are met, then it's considered accurate; otherwise, it's considered inaccurate.

\begin{table}[!ht]
  \centering
  \small
  \begin{tabular}{lccc}
    \toprule
    \textbf{Strategy Code} & \textbf{Identity} & \textbf{Rules} & \textbf{Candidates} \\
    \midrule
    \textbf{B (Baseline)}& English & English & English \\
    \textbf{C (Candidates)}& English & English & Source \\
    \textbf{I (Identity)}& Source & English & English \\
    \textbf{R (Rules)}& English & Source & English \\
    \textbf{I\&R (Identity \& Rules)}& Source & Source & English \\
    \textbf{A (All)}& Source & Source & Source \\
    \bottomrule
  \end{tabular}
  \caption{Translation strategies for prompt components across experimental configurations, showing which components remain in English versus being translated to the source language (Hindi or French).}\label{tab:translation-strategies}
\end{table}
\vspace{-20pt}

\begin{figure}[t]
  \begin{tcolorbox}[colback=gray!5, colframe=black, title=Example Prompt]
    \small
    \textcolor{identitycolor}{You are an intelligent AI agent and are assigned the task of intent detection based on speaker utterances. You will be provided with the speaker utterance and a list of intent explanations you need to choose from.}
    
    \vspace{0.5em}
    \textcolor{rulescolor}{You must follow these instructions:\\
    1) Carefully read the utterance and review all candidate intent explanations.\\
    2) Select the ONE intent explanation that most accurately represents the speaker utterance.\\
    3) Respond ONLY with the exact text of the chosen intent explanation. Do not modify the wording of the intent explanation.\\
    4) Do not add any explanations, greetings, or additional text.\\
    5) If multiple intents seem applicable, choose most specific one\\}
    
    \vspace{0.5em}
    \textcolor{candidatecolor}{\textbf{Candidate Intent Descriptions:}\\
    1. User wants to know the weather forecast for a specific location.\\
    2. User wants to set a reminder or alarm for a future event.\\
    3. User is requesting information about their account balance.\\
    4. User wants to play a specific song or music.\\
    5. User is asking for directions to a location.\\}
    
    \vspace{0.5em}
    \textcolor{utterancecolor}{\textbf{Utterance:}\\ "Peux-tu me rappeler de faire les courses à 17 heures ?"}
  \end{tcolorbox}
  \caption{The above figure presents the prompt we use for intent detection and its components: identity statement (blue), task rules (red), candidate intent descriptions (green), and user utterance (purple). The groundtruth response in this case would the utterance "User wants to set a reminder or alarm for a future event."}
  \label{fig:example-prompt}
\end{figure}

\subsection{Dataset}\label{sec:dataset}

\textit{Data Anonymization} Due to business considerations, we are not
permitted to share the results using the original user utterances. As a result, we manually anonymized both the labels and user utterances to ensure no personal information is included. Additionally, specific product and service names were denonymized to prevent the identification of the company from the user utterances or label descriptions. Despite these modifications, the conclusions drawn from our experiments remain valid.

The dataset comprises 6,000 utterances in French and in Hindi, respectively. We use 80 intent classes to categorize each user utterance. For each utterance, we select the candidate intents using embedding-based similarity, taking the top-k intents up to a certain threshold. To maintain full recall, we manually add the correct intent when it is not present among these top-k candidates.

\subsection{Cross-lingual Prompt Experiments}\label{prompts}
We specifically selected \textit{Hindi (Hi)} and \textit{French (Fr)} as the source languages for experimentation because they provide excellent diversity and represent different positions on the spectrum of language resource availability and linguistic similarity to English. Hindi is an under-represented language with lower resource availability and greater linguistic dissimilarity to English, while French is a high-resource language with relatively close linguistic similarities to English\cite{agrawal2024translationerrorssignificantlyimpact, cahyawijaya-etal-2024-llms}.

Throughout our experiments we keep the utterance (\textit{U}) in the source language and selectively translate a few other components of the prompt from English into the source language.  We use a LLM for translation and manually inspect a subset to ensure good translation quality. We prioritize component combinations that reflect authentic real-world usage scenarios. We name and abbreviate the experiment configurations based on what sections we translate into the source language. Table \ref{tab:translation-strategies} represents the set of translation components that we use. We treat the configuration with all components in English as \textbf{B (Baseline)} and compare the rest of the strategies against it. For multilingual RAG systems, knowledge bases are often maintained in English due to resource availability. Two common approaches emerge: (1) using cross-lingual retrieval\cite{zhuang2025multilingualinformationretrievalmonolingual} to find English documents for source language queries, resulting in English candidates, or (2) pre-translating the knowledge base into the source language and retrieving from the translated KB, resulting in source language candidates. Our Baseline configuration represents the first approach, while configuration \textbf{C (Candidates)} simulates the second approach where candidates are in the source language.

Additionally, configuration \textbf{I (Identity)} tests the impact of translating only the identity statement, providing insight into whether task framing in the source language affects comprehension. Configuration \textbf{R (Rules)} examines the effect of having instructions in the source language while keeping all other elements in English. The \textbf{I\&R (Identity \& Rules)} configuration combines both identity and rules translations to evaluate their synergistic effect, while still maintaining English candidates. Finally, configuration \textbf{A (All)} represents full localization, with all prompt components translated to match the utterance language. In configurations \textbf{C} and \textbf{A}, since the candidates are in the source language, the model response is also in the source language as the model just has to copy one of the candidates. We do not evaluate translating the user utterance into English, and instead primarily focus on methods where static prompt components can be pre-translated offline without impacting runtime performance.

\subsection{Models}\label{models}
For our experiments, we selected a diverse set of state-of-the-art multilingual LLMs to ensure comprehensive evaluation across different architectural designs and pre-training approaches. Specifically, we evaluated Llama-3.1-8B\cite{grattafiori2024llama3herdmodels}, Qwen2.5-7B-Instruct\cite{qwen2.5}, BLOOMZ-7b1\cite{muennighoff2022crosslingual}, Mistral-Nemo-Instruct-2407\cite{mistral_nemo_2024}, and BLOOMZ-7b1-mt\cite{muennighoff2022crosslingual}. This selection provides a balanced representation of both general-purpose models (Llama-3.1-8B) and those with enhanced instruction-following capabilities (Qwen2.5-7B-Instruct, Mistral-Nemo-Instruct-2407). Moreover, we deliberately included models with varying degrees of multilingual expertise, from those primarily trained on English (Llama-3.1-8B) to those specifically designed for multilingual tasks (BLOOMZ-7b1, BLOOMZ-7b1-mt). All models have comparable parameter sizes (approximately 7B-13B parameters range), allowing for fair comparison.

\begin{table*}[!ht]
  \centering
  \small
  \begin{tabular}{l|c|ccccc|c|ccccc}
    \toprule
    & \multicolumn{6}{c|}{\textbf{Hindi (Hi)}} & \multicolumn{6}{c}{\textbf{French (Fr)}} \\
    \cmidrule(lr){2-7} \cmidrule(lr){8-13}
    \textbf{Model} & \textbf{B} & \textbf{C} & \textbf{I} & \textbf{R} & \textbf{I\&R} & \textbf{A} & \textbf{B} & \textbf{C} & \textbf{I} & \textbf{R} & \textbf{I\&R} & \textbf{A} \\
    \midrule
    Llama-3.1-8B & 0.386 & \cellcolor{lightred}-4.9\% & \cellcolor{lightred}-0.8\% & \cellcolor{mediumred}-6.7\% & \cellcolor{mediumred}-7.3\% & \cellcolor{darkred}-10.9\% & 0.286 & \cellcolor{mediumred}-7.3\% & \cellcolor{darkred}-12.6\% & \cellcolor{darkred}-13.6\% & \cellcolor{darkred}-24.5\% & \cellcolor{darkred}-32.2\% \\
    Qwen2.5-7B-Instruct & 0.405 & \cellcolor{darkred}-12.8\% & \cellcolor{lightred}-3.2\% & \cellcolor{mediumgreen}+5.4\% & \cellcolor{mediumgreen}+6.7\% & \cellcolor{lightred}-2.2\% & 0.439 & \cellcolor{darkred}-12.5\% & \cellcolor{lightgreen}+3.4\% & \cellcolor{mediumgreen}+6.6\% & \cellcolor{mediumgreen}+5.7\% & \cellcolor{darkred}-13.4\% \\
    BLOOMZ-7b1 & 0.308 & \cellcolor{lightred}-1.9\% & \cellcolor{lightgreen}+0.3\% & \cellcolor{lightgreen}+3.2\% & \cellcolor{lightred}-3.2\% & \cellcolor{mediumred}-7.1\% & 0.186 & \cellcolor{mediumgreen}+6.5\% & \cellcolor{lightgreen}+2.2\% & \cellcolor{lightgreen}+1.1\% & \cellcolor{mediumgreen}+7.0\% & \cellcolor{darkgreen}+13.4\% \\
    Mistral-Nemo-Instruct-2407 & 0.454 & \cellcolor{darkred}-13.0\% & \cellcolor{lightred}-2.9\% & \cellcolor{darkred}-15.0\% & \cellcolor{darkred}-16.7\% & \cellcolor{darkred}-33.3\% & 0.443 & \cellcolor{lightred}-2.3\% & \cellcolor{lightgreen}+0.5\% & \cellcolor{lightred}-2.3\% & \cellcolor{lightred}-2.3\% & \cellcolor{mediumred}-9.0\% \\
    BLOOMZ-7b1-mt & 0.308 & \cellcolor{lightred}-0.3\% & \cellcolor{lightgreen}+1.9\% & \cellcolor{lightgreen}+2.3\% & \cellcolor{lightgreen}+1.0\% & \cellcolor{lightred}-1.3\% & 0.185 & \cellcolor{mediumgreen}+7.0\% & \cellcolor{lightgreen}+1.6\% & \cellcolor{lightred}-1.6\% & \cellcolor{lightgreen}+4.3\% & \cellcolor{mediumgreen}+9.7\% \\
    \bottomrule
  \end{tabular}
  \caption{Performance comparison of different prompt translation strategies across languages (Hindi and French). Values for baseline (B) are absolute accuracy scores, while other columns (C, I, R, I\&R, A) show percentage changes relative to the baseline. \textcolor{darkgreen}{Dark green} cells indicate strong improvements ($\geq$10\%), \textcolor{mediumgreen}{medium green} for moderate improvements (5-10\%), and \textcolor{lightgreen}{light green} for small improvements (0-5\%). Similarly, \textcolor{darkred}{dark red} shows strong declines ($\geq$10\%), \textcolor{mediumred}{medium red} for moderate declines (5-10\%), and \textcolor{lightred}{light red} for small declines (0-5\%).}
  \label{tab:cross-lingual-performance}
\end{table*}

\subsection{Implementational Details}
As mentioned earlier, we evaluate these models with various prompt configurations without explicit tuning on the dataset.  We use standard inference parameters of temperature as 0.1, top-p as 0.7 and max\_tokens as 512 to have deterministic generation properties. We perform distributed inference using vllm\cite{kwon2023efficient} on a machine with 4 NVIDIA L4 GPUs. 

\section{Results}\label{sec:exp}

We compute the average accuracy across all the prompt configurations and present the results in Table \ref{tab:cross-lingual-performance}. Given the baseline performance we report the percentage improvement or decrement in accuracy for each model and prompt configuration.

We observe relatively low absolute baseline performance across all models for the intent detection task. We attribute this to two main factors: i) The intent detection instances are out-of-domain (OOD) for the LLMs, as they have not encountered them during training, and ii) LLMs struggle with verbatim copying portions of the user prompt \cite{wang-etal-2024-positionid}. Furthermore, we observe varying trends across different LLMs for each prompt configuration. Specifically, Llama-3.1-8B and Mistral-Nemo-Instruct-2407 experience performance decreases when any component is translated into the source language. In contrast, the BLOOMZ-based models show decent improvements in performance across most prompt configurations. Qwen2.5-7B-Instruct demonstrate mixed results, with improvements in the \textbf{I}, \textbf{R} and \textbf{I\&R} strategies while decrements in \textbf{C} and \textbf{A}. From a language perspective, we notice that all models struggled with generating Hindi text (in configurations \textbf{C} \& \textbf{A}), experiencing a decline in performance, while for French, generation in the native language proves beneficial, but only for the BLOOMZ models. In general, having Identity, Rules, or both in the respective language leads to some performance improvements.

\section{Discussion and Recommendations}
We hypothesize and discuss the following insights about choosing the optimal prompt language for a task:

\paragraph{\textbf{LLM generation abilities are better in English, especially for low resource languages.}}Most models perform poorly when the candidates are presented in the source language, thereby forcing the model to generate the response in source language. The only exception is that for the BLOOMZ models in the case of the French language. Such observation is expected and can be attributed to the fact that the amount of training data in English is disproportionately large as compared to other languages. Similar observations have been made in several other works\cite{lai-etal-2023-chatgpt, wu2025semantichubhypothesislanguage}.

\paragraph{\textbf{Keeping Identity and Instructions in the source language can lead to improvements.}}
Our results indicate that translating the identity statement (\textbf{I}) and rules (\textbf{R}) into the source language often yields performance improvements, particularly for models like Qwen2.5-7B-Instruct and the BLOOMZ variants. We hypothesize that framing the task in the user's native language helps the model better understand the context and intent of the query, while still leveraging the model's stronger English generation capabilities. This suggests that a hybrid approach—where task framing occurs in the source language but generation remains in English—may be optimal for many multilingual applications.
\paragraph{\textbf{Model architecture and pre-training strategy significantly impact cross-lingual performance.}} The varying performance across different models suggests that architectural design and pre-training objectives play crucial roles in determining cross-lingual capabilities. For instance, the BLOOMZ models, which were specifically trained with multilingual objectives, demonstrate more consistent improvements across different prompt configurations compared to the others. This indicates that the choice of LLM should be carefully considered based on the specific language requirements of the application.
\paragraph{\textbf{Language resource availability and task requirements drive optimal translation strategy.}} Our results reveal contrasting optimal strategies between Hindi (low-resource) and French (high-resource). Hindi generation showed performance degradation across most models, while French generation benefited certain models, suggesting that language resource availability during pre-training significantly impacts prompt translation effectiveness. The inconsistent performance across configurations also indicates that optimal strategies depend on specific task requirements—for intent classification, translating identity and rules proved beneficial but may not generalize to other tasks. This highlights the need for optimization based on both language resource levels and task-specific demands.
\vspace{-1pt}
\section{Conclusion and Future Work}
This work addresses a fundamental challenge in multilingual dialogue systems: optimizing cross-lingual prompt strategies without sacrificing runtime performance.
Throughout this study, we systematically evaluate the impact of translating parts of the system prompt into the source language of dialogue queries for the dialogue intent detection task. Our results show that the choice of optimal prompt strategy can vary greatly based on the type of language and the inference model being used. In general, we observed improvements when the identity and instructions of the prompt were translated to the source language. For low-resource languages that are lexically dissimilar to English, like Hindi, it's always better to have the generation in English unless the task requires Hindi text. For high-resource languages like French, it can be better to have generation in the source language, depending on the choice of the inference model. Different LLMs behave differently when presented with cross-lingual prompts, which can be attributed to the different data and recipes used to train these models. While models like BLOOMZ-7b1 and BLOOMZ-7b1-mt show consistent improvements in performance with translated prompts, others like Llama-3.1-8B perform better with English-only prompts.

For future work, we plan to extend our analysis to a broader range of language tasks beyond intent classification, including question answering, summarization, and entity extraction, to determine whether our hypotheses about translation strategies generalize across diverse task types. Additionally, we aim to expand our language coverage to include more typologically diverse languages representing various resource levels to better understand how linguistic distance from English impacts optimal prompt translation strategies. We also intend to investigate the role of translation quality more systematically by comparing professional human translations against various machine translation approaches, as well as exploring hybrid prompting strategies that dynamically select which components to translate based on model confidence scores.

\bibliographystyle{ACM-Reference-Format}
\bibliography{sample-base}

\end{document}